\documentclass[10pt,twocolumn,letterpaper]{article}

\usepackage[pagenumbers]{wacv} 

\usepackage{amsmath,amssymb}
\usepackage{booktabs}
\usepackage{multirow}
\usepackage{makecell}

\makeatletter
\renewcommand\paragraph{\@startsection{paragraph}{4}{\z@}%
  {4pt plus 1pt minus 1pt}{-1em}{\normalsize\bfseries}}
\makeatother

\AtBeginDocument{%
  \setlength{\abovedisplayskip}{4pt plus 1pt minus 1pt}%
  \setlength{\belowdisplayskip}{4pt plus 1pt minus 1pt}%
  \setlength{\abovedisplayshortskip}{2pt plus 1pt}%
  \setlength{\belowdisplayshortskip}{2pt plus 1pt}%
}

\makeatletter
\def\wacvsection{\@startsection {section}{1}{\z@}
   {-8pt plus -2pt minus -2pt}{5pt} {\large\bf}}
\def\wacvsubsection{\@startsection {subsection}{2}{\z@}
   {-6pt plus -2pt minus -2pt}{3pt} {\elvbf}}
\makeatother

\definecolor{wacvblue}{rgb}{0.21,0.49,0.74}
\usepackage[pagebackref,breaklinks,colorlinks,allcolors=wacvblue]{hyperref}

\usepackage[capitalize]{cleveref}
\crefname{section}{Sec.}{Secs.}
\Crefname{section}{Section}{Sections}
\crefname{table}{Tab.}{Tabs.}
\Crefname{table}{Table}{Tables}
\newcommand{\suppref}[1]{\cref{#1}}
\newcommand{\Suppref}[1]{\Cref{#1}}

\def\wacvPaperID{2732}
\def\confName{WACV}
\def\confYear{2027}

\title{Thinking from Above:\\Efficient Spatial Reasoning with Bird's-Eye-View Visual Tokens}

\author{Faizan Siddiqui \qquad Nirav Savaliya \qquad Sahith Reddy Chada \qquad Enna Sachdeva\\
Honda Research Institute USA\\
{\tt\small faizan\_siddiqui@honda-ri.com}
}

\begin{document}
\maketitle
\begin{abstract}
Existing video multimodal large language models (MLLMs) organize observations primarily in time, but embodied spatial reasoning requires organizing them in space. Consequently, their visual representations can remain locally consistent across nearby frames while lacking a persistent spatial context that is stable across viewpoints and time. We exploit depth and camera pose, which are naturally available to embodied agents, to explicitly establish this spatial correspondence. We introduce Cambrian-BEV, which extends Cambrian-S with a parameter-free geometric tokenization scheme that back-projects visual patch features from each frame into a shared bird's-eye-view (BEV) coordinate frame and averages features that fall within the same spatial cell. Thus, observations of the same physical location, regardless of viewpoint or time step, are aggregated into a common spatial token. In addition to providing a persistent scene representation, this formulation acts as an efficient compression mechanism: 128 frames containing approximately 100K DINOv3 patch features are consolidated into a fixed 27$\times$27 grid of only 729 BEV tokens, cutting LLM prefill time by $2.9\times$ and peak memory by $3.6\times$. Cambrian-BEV requires no architectural modification beyond BEV tokenization and uses a frozen DINOv3 vision encoder with a trainable multimodal projector and LLM. On VSI-Bench, our model reaches 63.48, $+8.5$ points over an identically fine-tuned frame-token control with $12.6\times$ fewer visual tokens, with particularly large gains on room size ($+24.5$) and relative direction ($+34.1$). On ReVSI, it reaches 44.68, $+4.5$ over the corresponding control.
\end{abstract}
\section{Introduction}
\label{sec:intro}

\begin{figure*}
\centering
\includegraphics[width=\linewidth]{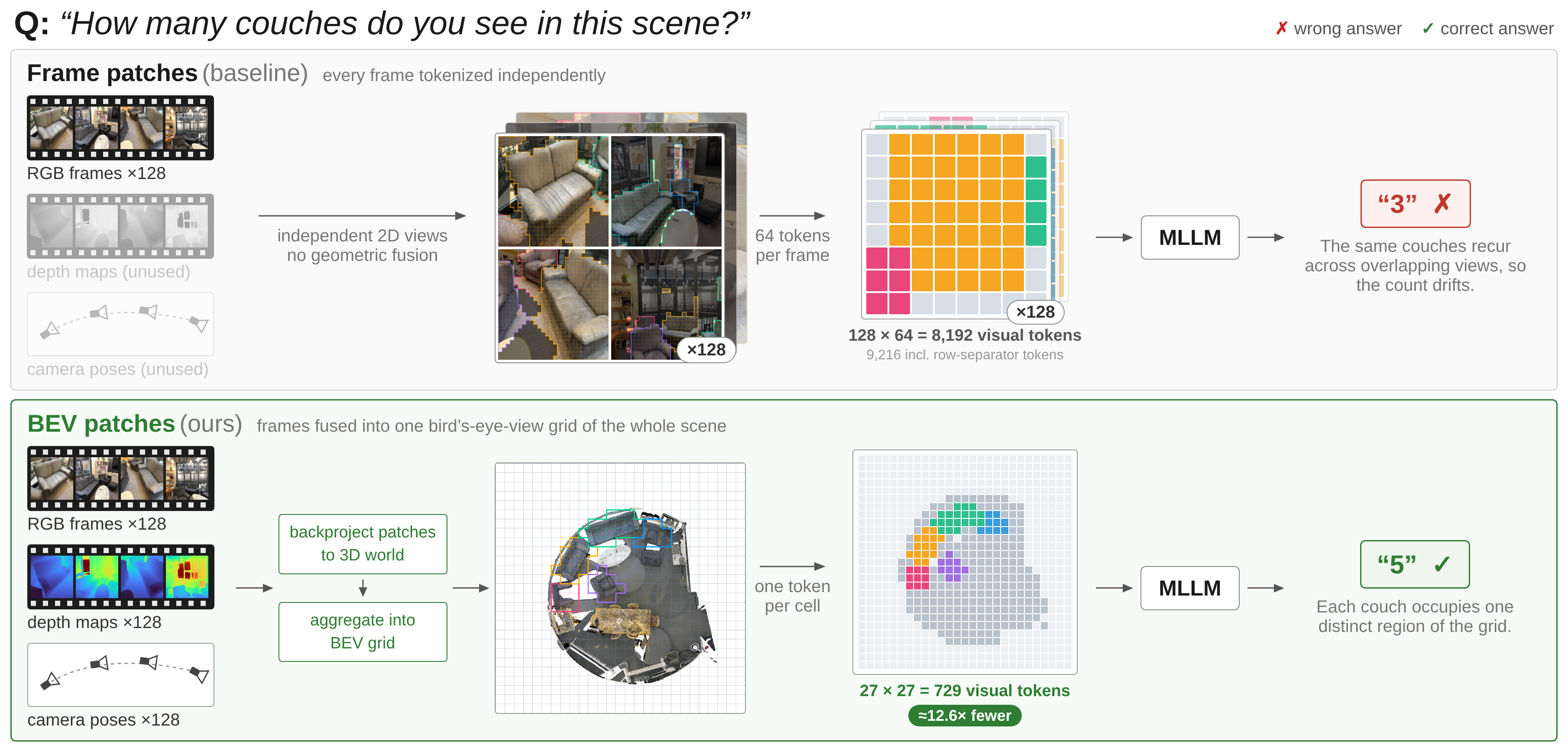}
\caption{\textbf{From frame-indexed tokens to BEV visual tokens.}
\emph{Top:} A conventional video MLLM represents each frame independently, producing repeated evidence for regions observed from multiple viewpoints.
\emph{Bottom:} We use depth and camera pose to back-project patch features into a shared metric coordinate frame and aggregate them on a fixed $27 \times 27$ bird's-eye-view grid. Repeated observations of the same spatial region are fused into common BEV cells, replacing approximately 9K frame tokens with 729 spatially indexed tokens at the LLM interface.}
\label{fig:overview-figure}
\end{figure*}

Multimodal large language models (MLLMs) have become highly capable at recognizing and describing visual content, yet they remain substantially weaker at reasoning about the geometry of a scene. In video-based spatial question answering, a scene is typically represented as a sequence of frame-level visual tokens~\cite{liu2023visual,tong2024cambrian,yang2026cambrians} (\cref{fig:overview-figure}, top). Each token is indexed by the image coordinates of a particular viewpoint, rather than by the physical location it observes. Consequently, overlapping views produce redundant evidence, while the language model must implicitly establish cross-view correspondence, reconcile revisited regions, and infer metric relationships from a long token sequence. VSI-Bench highlights this limitation: although local spatial awareness emerges in modern MLLMs, constructing and reasoning over a coherent scene-level representation remains a major bottleneck~\cite{yang2025thinking}. Increasing the number of frames can provide broader visual coverage, but also increases the visual context approximately linearly, creating a particularly unfavorable trade-off for compact models intended for embodied deployment.

This trade-off is unnecessary in many embodied settings. Robots, augmented-reality devices, and mobile scanning platforms commonly provide, or can estimate, depth, camera intrinsics, and camera pose. These signals are also available in indoor datasets such as ScanNet, ScanNet++, and ARKitScenes~\cite{dai2017scannet,yeshwanth2023scannetpp,baruch2021arkitscenes}. Given this information, an image patch can be associated with a location in a common 3D coordinate frame without learning a view transformation. Spatial mapping and BEV perception have long used this principle to align observations across viewpoints~\cite{cartillier2021semanticmapnet,philion2020lift,harley2023simplebev}. Recent spatial MLLMs have also begun to incorporate camera pose, reconstructed geometry, BEV renderings, or explicit 3D memories~\cite{zheng2025video3dllm,fan2026vlm3r,yang2026cambrianp,qi2026gpt4scene,gwak2026cog3dmap}. However, these methods often retain frame-indexed visual tokens, depend on learned geometric modules, render object-level maps, or maintain variable-length spatial memories. We instead ask whether known geometry can be used directly as a visual tokenization rule, converting a long video into a fixed set of scene-indexed features before language reasoning begins, so the model reasons over the scene from above rather than frame by frame.

We build on Cambrian-S~\cite{yang2026cambrians}, a video MLLM whose spatial instruction tuning on the VSI-590K corpus improves spatial reasoning but leaves this frame-indexed interface in place. We introduce \textbf{Cambrian-BEV}, which keeps the Cambrian-S architecture, data, and training recipe and replaces only the visual interface with a parameter-free geometric one: a compact bird's-eye-view representation in place of frame tokens (\cref{fig:overview-figure}, bottom). Using depth, camera intrinsics, and pose, we back-project dense patch features from a frozen DINOv3 encoder into a shared scene coordinate frame, average them within a fixed $27 \times 27$ BEV grid, and supply the resulting 729 scene-indexed tokens to the LLM in place of the frame-token sequence. For a 128-frame input, this reduces the visual context presented to the LLM by approximately $12.6\times$ relative to our frame-token baseline. The BEV operator introduces no learnable parameters and requires no object detector or map supervision; only the multimodal projector and LLM are fine-tuned, on the depth-and-pose subset of VSI-590K.

Geometric rearrangement also imposes a requirement that conventional video MLLMs hide: a patch feature must remain spatially tied to the image region from which it was extracted, a property that frame-token performance cannot reveal because tokens are normally consumed in their original image order. Comparing SigLIP2~\cite{tschannen2025siglip2}, the vision encoder used by Cambrian-S, with DINOv3~\cite{simeoni2025dinov3}, we find that the two perform nearly identically as frame-token models but diverge substantially once their patch features are rearranged in BEV space, making dense feature locality a key, and previously underexamined, criterion for selecting vision encoders for geometry-aware MLLMs (\cref{sec:method-encoder}).

On VSI-Bench, Cambrian-BEV with a compact 0.5B-parameter language backbone reaches 63.48, improving by $8.50$ points over an identically fine-tuned frame-token control while using $12.6\times$ fewer LLM-side visual tokens, which cuts LLM prefill time $2.9\times$ and peak memory $3.6\times$. The largest gains occur on room size, relative direction, and route planning; the metric gains hold across every BEV configuration we trained, whereas the relative-direction gain requires anchoring the grid to the first camera's heading. The principal trends replicate on the independently re-annotated ReVSI benchmark~\cite{zhang2026revsi}, where Cambrian-BEV improves over the corresponding control by $4.51$ points.

Our contributions are threefold:
\begin{itemize}
\item \textbf{A parameter-free BEV tokenization interface for video MLLMs.}
We use depth and camera pose to co-register dense patch features in a fixed metric grid, replacing a long frame-token sequence with 729 scene-indexed visual tokens and reducing the LLM visual context by approximately $12.6\times$.

\item \textbf{An empirical study of encoder locality under geometric rearrangement.}
We show that comparable frame-token performance does not imply comparable BEV performance, identifying spatially localized dense features as a key requirement for geometry-aware visual tokenization.

\item \textbf{Strong spatial reasoning with a compact language backbone.}
Cambrian-BEV reaches 63.48 on VSI-Bench and 44.68 on ReVSI, with robust improvements on metric questions (room size, distances) and, given an egocentric grid frame, on relative direction, demonstrating the value of explicit geometric alignment for efficient embodied video understanding.
\end{itemize}
\section{Related Work}
\subsection{Spatial Reasoning in Video MLLMs}

Spatial question answering initially focused on benchmarks and models that operate on explicit 3D scene representations~\cite{azuma2022scanqa,ma2023sqa3d,zhu20233dvista,hong20233dllm}. Subsequent methods introduced spatial awareness into 2D vision-language models through spatial instruction tuning, depth prediction, and RGB-D conditioning~\cite{chen2024spatialvlm,cheng2024spatialrgpt,cai2025spatialbot}. More recently, video-based benchmarks such as VSI-Bench and ReVSI have shown that modern MLLMs remain particularly limited in metric, configurational, and cross-view reasoning~\cite{yang2025thinking,zhang2026revsi}. Geometry-aware video MLLMs address these limitations using 3D positional encodings, pretrained reconstruction features, spatial instruction tuning, pose-grounded camera tokens, or combinations of implicit and explicit geometry~\cite{zheng2025video3dllm,zheng2025vgllm,fan2026vlm3r,yang2026cambrians,yang2026cambrianp,li2026vlmie3d,huang2026consispace}. Despite substantial progress, many of these approaches retain observations as frame- or view-indexed tokens and incorporate geometry as auxiliary information. Repeated observations of the same physical region can therefore remain represented by separate tokens, leaving the language model to infer their correspondence. Our approach instead uses available depth and camera pose to establish this correspondence before language reasoning by organizing visual features in a shared spatial coordinate frame.

\subsection{Allocentric and BEV Scene Representations}

Allocentric representations provide a persistent coordinate system in which observations from different viewpoints can be spatially aligned. Semantic MapNet and Trans4Map project egocentric visual features into top-down memories for semantic-map prediction~\cite{cartillier2021semanticmapnet,chen2023trans4map}, while VLMaps and ConceptFusion associate geometric maps with open-vocabulary visual-language features for navigation and multimodal retrieval~\cite{huang2023vlmaps,jatavallabhula2023conceptfusion}. Recent work increasingly exposes structured allocentric representations to multimodal language models. GPT4Scene renders a BEV map and uses consistent object identifiers to connect top-down and egocentric observations~\cite{qi2026gpt4scene}; Struct2D combines BEV renderings with object-level metadata~\cite{zhu2025struct2d}; and Video2Layout and Map2Thought reconstruct metric layouts that support explicit spatial computation~\cite{huang2025video2layout,gao2026map2thought}. Cog3DMap instead constructs a variable-length 3D token memory, recurrently merging semantic and geometric features associated with overlapping spatial locations~\cite{gwak2026cog3dmap}. These methods demonstrate the value of allocentric representations, but typically render symbolic maps, reconstruct object-level layouts, use learned mapping modules, or maintain variable-size 3D memories. In contrast, we directly back-project dense visual patch features into a fixed metric BEV grid and average all features assigned to each cell. The geometric projection and aggregation are parameter-free and require no object detector, semantic-map supervision, map decoder, or learned view transformation. The resulting BEV cells are supplied to the MLLM as spatially aligned visual tokens.

\subsection{Token-Efficient Video Modeling}

The visual context of a video MLLM grows rapidly with the number of processed frames, motivating representations that preserve long-range evidence under a bounded token budget. Prior approaches maintain compact working and episodic memories~\cite{liu2026vlm2}, fold streaming observations into persistent memory structures~\cite{wei2026streaming}, or incrementally compress geometry-aware features through spatial clustering or voxelization~\cite{yu2026stream3dvlm,gwak2026cog3dmap}. Structured scene graphs and 3D memories provide another form of compression by retaining object-level entities and relations rather than dense frame features~\cite{gu2024conceptgraphs,zemskova2025graphllm,yang20253dmem}. Such methods generally reduce context through learned memory updates, selective retention, object abstraction, or adaptive spatial grouping. Our method instead uses deterministic geometric correspondence as the compression criterion. Features are merged when their back-projected locations fall within the same physical BEV cell, regardless of the frame or viewpoint from which they were observed. Consequently, the token budget is bounded by the BEV grid resolution rather than video length, making spatial alignment and video compression two outcomes of the same operation.
\section{Methodology}
\label{sec:methodology}

\begin{figure*}
  \centering
  \includegraphics[width=\linewidth]{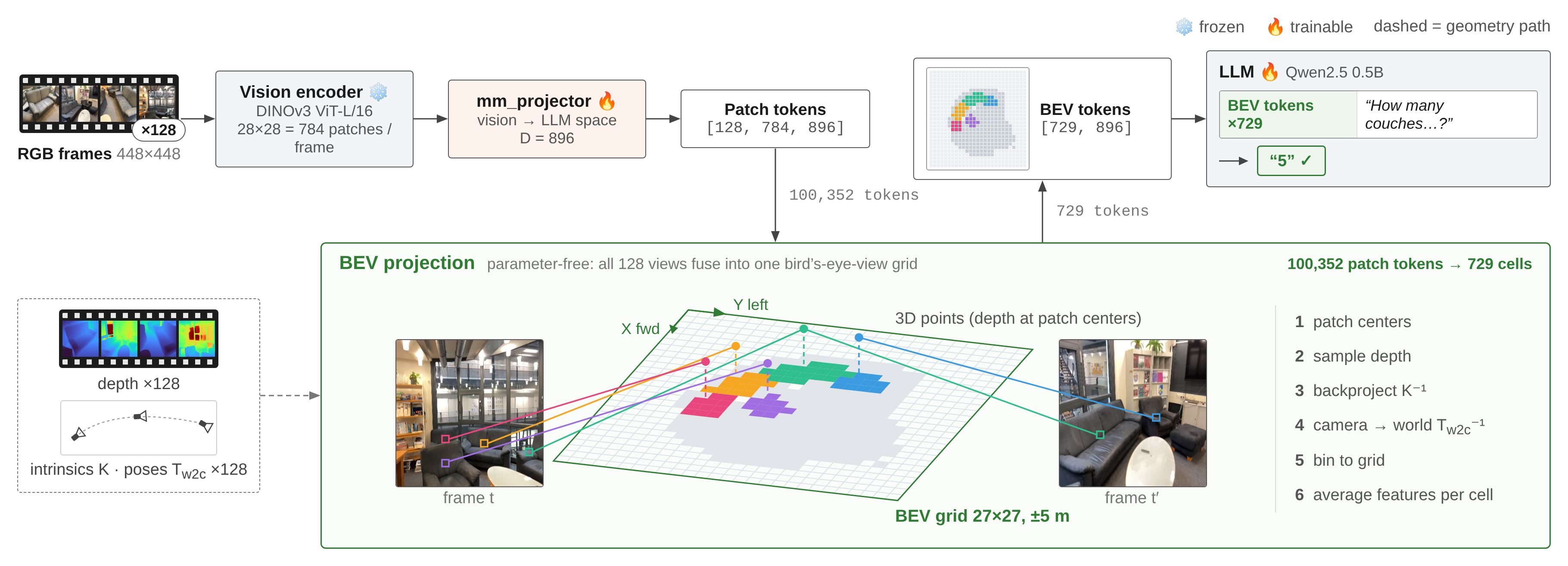}
  \caption{\textbf{Cambrian-BEV architecture.} Patch tokens leave the frozen encoder and the vision-to-language projector already in the language embedding space, so the parameter-free geometric module (steps 1--6) can pool them directly into a metric BEV grid whose cells become the BEV tokens that replace the frame tokens in the LLM context. The BEV projection (green) is introduced only in the final fine-tuning stage (stage~5, \suppref{tab:supp-stages}); the encoder, projector, and LLM are inherited from the unmodified four-stage Cambrian-S recipe, which trains on frame tokens.}
  \label{fig:architecture-figure}
\end{figure*}

Our goal is to replace the frame-sequential visual context of a video MLLM with a compact, metrically structured one while modifying the model as little as possible. \Cref{fig:architecture-figure} shows the pipeline. The base model's frozen encoder and projector produce patch tokens (\cref{sec:method-prelim}); a parameter-free geometric module fuses them across frames into a single BEV grid (\cref{sec:method-projection}); and the LLM consumes this grid in place of image tokens (\cref{sec:method-integration}). \Cref{sec:method-encoder} analyzes the encoder property that makes this fusion viable.

\subsection{Preliminaries}
\label{sec:method-prelim}

We build on Cambrian-S~\cite{yang2026cambrians}, a video MLLM comprising a vision encoder, a two-layer MLP projector into the language embedding space, and a Qwen2.5-0.5B language model~\cite{qwen2024qwen25} with hidden size 896. Cambrian-S is trained in four stages: (1)~projector alignment on image--caption pairs, (2)~image instruction tuning, (3)~video instruction tuning, and (4)~spatial instruction tuning on VSI-590K. Only stage~1 restricts training to the projector; the LLM is trained thereafter, and the vision encoder stays frozen throughout (\suppref{tab:supp-stages}). The input is a scene video of $T{=}128$ RGB-D frames with per-frame camera intrinsics $K_t \in \mathbb{R}^{3\times3}$ and world-to-camera extrinsics $T^{w2c}_t \in SE(3)$. These quantities are natively available on embodied platforms and are provided by the ScanNet~\cite{dai2017scannet}, ScanNet++~\cite{yeshwanth2023scannetpp}, and ARKitScenes~\cite{baruch2021arkitscenes} scenes underlying VSI-590K and VSI-Bench. Each frame is encoded at $448{\times}448$ by a frozen DINOv3~\cite{simeoni2025dinov3} ViT-L/16 into a $28{\times}28$ grid of patch features, which the projector maps to language-model width. This yields patch tokens $F_t \in \mathbb{R}^{N_p \times D}$ with $N_p{=}784$ and $D{=}896$ per frame. The baseline model compresses these $T \cdot N_p \approx 100$K features per scene: Cambrian-S's multi-image/video (MIV) pathway bilinearly downsamples each $28{\times}28$ grid to $8{\times}8$ tokens, appends a learnable newline token per row, and concatenates the 72 tokens per frame in time, giving $9{,}216$ visual tokens. We discard that pathway entirely and instead route the full-resolution patch tokens through the BEV module.

\subsection{Parameter-free BEV projection}
\label{sec:method-projection}

The BEV module assigns each patch token a 3D location by closed-form geometry and then pools tokens by ground-plane cell. All geometric computation is deterministic given depth and pose, and nothing is learned. The module is differentiable with respect to the input features, so gradients from the language loss backpropagate through the pooling operation to the projector (\suppref{sec:supp-projection}). Because pooling operates on \emph{projector outputs}, the resulting BEV tokens already live in the language embedding space, and no adapter of any kind is required (\cref{sec:method-integration}).

\paragraph{Back-projection.}
The patch at row $i$, column $j$ of frame $t$ has center pixel $(u, v) = \big((j{+}\tfrac12)P,\, (i{+}\tfrac12)P\big)$ for patch size $P$. We sample the depth map bilinearly at this point to obtain $z^{cam}$, discarding patches whose depth is invalid or outside $[0.1, 100]$\,m. The pinhole model lifts the patch center into the camera frame,
\begin{equation}
  \mathbf{p}^{cam} = z^{cam} \, K_t^{-1} \, (u, v, 1)^\top,
\end{equation}
and the inverse extrinsics carry it to the world frame:
\begin{equation}
  \mathbf{p}^{w} = R_t^\top \big( \mathbf{p}^{cam} - \mathbf{t}_t \big), \qquad
  T^{w2c}_t = \begin{bmatrix} R_t & \mathbf{t}_t \\ \mathbf{0} & 1 \end{bmatrix}.
\end{equation}

\paragraph{Grid orientation and binning.}
The BEV grid is an $H_b {\times} W_b$ Cartesian lattice on the ground plane, with rows spanning the forward axis and columns the lateral axis. World-frame axes are arbitrary with respect to both the room and the observer, and the choice of grid frame matters a great deal. We therefore rotate ground-plane coordinates so that the grid's forward axis aligns with the first camera's heading (its optical axis projected onto the ground plane). This gives the grid an egocentric, trajectory-anchored meaning: ``forward as the video first saw it.'' This choice of frame turns out to be the largest positive lever in our ablations (\cref{sec:exp-ablations}). The grid is centered on the centroid of the valid points and spans $[-5, 5]$\,m in both axes at $H_b{=}W_b{=}27$ (0.37\,m per cell). The range is set from the training scenes' spatial extent rather than tuned: a centered $\pm5$\,m window retains 99\% of back-projected points, and smaller windows discard scene content faster than they gain resolution (\suppref{sec:supp-grid-range}). Each surviving point is assigned to the cell containing its ground-plane coordinates. \Suppref{sec:supp-projection} specifies the frame conventions, the PCA-rectified alternative, the centering and binning rules, and the degenerate cases in full.

\paragraph{Aggregation.}
Each surviving point $p$ deposits its patch feature $f_p$ into the cell containing its ground-plane coordinates, and every cell averages the features it receives:
\begin{equation}
  B_{ij} =
  \begin{cases}
    \dfrac{1}{|\mathcal{P}_{ij}|} \displaystyle\sum_{p \in \mathcal{P}_{ij}} f_p, & |\mathcal{P}_{ij}| > 0,\\[1.4ex]
    \mathbf{0}, & |\mathcal{P}_{ij}| = 0,
  \end{cases}
  \label{eq:aggregation}
\end{equation}
where $\mathcal{P}_{ij}$ is the set of points assigned to cell $(i, j)$ and $\mathbf{0} \in \mathbb{R}^{D}$ is the zero vector. Cells that receive no points are therefore exactly zero rather than undefined or imputed, and these zero tokens are passed to the LLM unchanged (\suppref{sec:supp-empty-cells}). We implement this as a single batched \texttt{scatter\_add} followed by a count normalization. On an H100, the full projection of a 128-frame scene ($128{\times}784$ points), from depth sampling to the normalized grid, takes 1.9\,ms (\suppref{sec:supp-runtime}). We also evaluate a bilinear variant that splats each point onto its four surrounding cells with bilinear weights and normalizes by accumulated weight, with the same zero-vector convention for cells whose accumulated weight is zero; we compare the two in \cref{sec:exp-ablations} and \cref{sec:exp-revsi}. The per-cell averaging of \cref{eq:aggregation} is where multi-view fusion happens. The same physical location observed from ten viewpoints contributes ten features to one cell, so redundant observations collapse into a stable estimate rather than ten near-duplicate tokens.

\paragraph{Vertical structure.}
Collapsing the height axis means that a cell can mix an object with the wall, floor, and ceiling behind it. Measured on our final $27{\times}27$, $\pm5$\,m grid, an object in the 0.3--1.2\,m height band typically shares its cell with 1.5--2.7$\times$ its own feature mass, and up to 169$\times$ in the worst case. We therefore support an optional per-scene floor-relative height filter (\suppref{sec:supp-height-filter}). Whether removing the diluting mass helps is an empirical question; it does not (\cref{tab:ablations}e, \suppref{sec:supp-ablations}), and the final model uses no height filter.

The output is a grid $B \in \mathbb{R}^{H_b \times W_b \times D}$, flattened row-major into $H_b W_b = 729$ BEV tokens. This is a fixed-size, scene-indexed set of BEV tokens, $12.6\times$ smaller than the frame-token context it replaces, in which neighboring tokens are metric neighbors in the world.

\subsection{Fusion needs locally grounded features}
\label{sec:method-encoder}

BEV aggregation implicitly assumes that a patch token describes the physical surface at that patch. If each token is instead a mixture of scene-global context, then rearranging tokens by 3D position is meaningless, and averaging tokens from different views of the same place blurs whatever structure they carried. Whether the assumption holds is a property of the encoder's training objective.

Language-supervised encoders such as SigLIP2~\cite{tschannen2025siglip2} receive their contrastive signal through an attention-pooling head over patch tokens. The objective constrains only the pooled global embedding, leaving individual patch tokens free to distribute evidence across the image. Probing SigLIP2 directly confirms this (\cref{fig:text-sim}). We read its patch tokens in the way most favorable to it, carrying each through the value pathway of the pooling head so that it becomes the per-patch analogue of the embedding the text tower is trained against; the raw pre-head tokens are less localized still (\suppref{sec:supp-text-sim}). Even so, the resulting text-similarity maps scatter evidence for a queried object across the image, whereas those of DINOv3 follow the object's extent. DINOv3~\cite{simeoni2025dinov3}, in contrast, is trained with dense patch-level self-distillation and adds \emph{Gram anchoring}, a regularizer that preserves the pairwise similarity structure of patch features throughout training. Its features are therefore designed to preserve stronger dense feature structure.

This leads to a testable prediction. If BEV fusion depends on feature locality, then two encoders that perform equally as conventional frame-token models should diverge once their tokens are geometrically fused, with the locally grounded encoder benefiting more. We test this by training both encoders through the identical four-stage recipe (\suppref{tab:supp-stages}) followed by the identical BEV fine-tuning stage (\cref{sec:exp-ablations}).

\subsection{Integration into the language model}
\label{sec:method-integration}

The 729 BEV tokens are spliced into the embedded sequence at the position of the \texttt{<image>} placeholder, exactly where the frame tokens would have gone. The LLM architecture, tokenizer, and prompt format are untouched. The final model introduces no new BEV-specific parameters, BEV tokens are already language-embedding vectors, and an interposed MLP adapter is not only unnecessary but harmful ($-12.0$ points, \cref{tab:ablations}d). We also evaluate richer interfaces, namely a learnable empty-cell token, a small self-attention BEV encoder with axial 2D rotary embeddings, and temporally chunked (iterative) BEV sequences. None improves on the plain parameter-free grid (\cref{tab:ablations}, \suppref{sec:supp-ablations}).

\paragraph{Training.}
Starting from the stage-4 (spatial IT) checkpoint, we fine-tune the projector and LLM with the vision encoder frozen, using the standard next-token loss on answer tokens only. Because the BEV grid is a per-scene quantity, training batches are scene-grouped so that the vision encoding and projection are computed once per scene and shared across its QA pairs. Hyperparameters are given in \cref{sec:exp-setup}.

\section{Experiments}
\label{sec:experiments}

\subsection{Setup}
\label{sec:exp-setup}

\paragraph{Training data.}
Models are fine-tuned on the subset of VSI-590K~\cite{yang2026cambrians} whose scenes provide registered depth and pose, which we refer to as \textbf{VSI-288K-Depth}: 288{,}478 QA pairs over 4{,}955 scenes (1{,}201 ScanNet, 855 ScanNet++, 2{,}899 ARKitScenes), with 128 uniformly sampled frames per scene. Depth is raw sensor depth (16-bit, millimeter) and poses are the datasets' ground-truth trajectories; we perform no reconstruction or pose estimation. VSI-590K covers counting, distance, size, direction, and appearance order but contains no route-planning questions. Route planning, one of the eight VSI-Bench categories, is therefore a held-out task for every fine-tuned model below.

\paragraph{Models.}
All 0.5B rows in \cref{tab:main-results} share the same architecture and the same four-stage Cambrian-S recipe (\cref{sec:method-prelim}, \suppref{tab:supp-stages}); they differ only in the vision encoder and in the visual representation presented to the LLM.
\emph{Frames (SigLIP2)} is our reproduction of Cambrian-S-0.5B with its original SigLIP2-So400M encoder.
\emph{Frames (DINOv3)} replaces the encoder with DINOv3 ViT-L/16 and serves as the parent checkpoint of every BEV model below.
\emph{Cambrian-BEV} continues this parent with a fifth fine-tuning stage on VSI-288K-Depth in which the $9{,}216$ frame tokens (128 frames $\times$ 72 MIV tokens, \cref{sec:method-prelim}) are replaced by 729 BEV tokens ($27{\times}27$ grid, $\pm5$\,m range, hard assignment, first-pose orientation, no adapter).
\emph{Frames + spatial FT} is the critical control. It gives the same DINOv3 parent the \emph{identical} fifth stage on the identical data but retains its $9{,}216$ frame tokens, so the gap between this row and ours isolates the effect of the representation from the effect of the extra training. For reference, we also report the official Cambrian-S releases at 0.5B and 7B, evaluated in our harness.

\paragraph{Implementation details.}
The fifth stage shared by Cambrian-BEV and the control runs for 2 epochs (2{,}176 steps) with the projector and LLM fully trainable (no LoRA), both at learning rate $2{\times}10^{-5}$ with a cosine schedule, 3\% warmup, and no weight decay. The global batch is 256, organized as 8 QA pairs per scene. As in stages 1--4, training runs under bf16 autocast with fp32 master weights (\suppref{sec:supp-repro-gap}). Training takes about 6 hours on 2$\times$H100 80\,GB GPUs.

\paragraph{Benchmarks and metrics.}
Our primary benchmark is the full VSI-Bench~\cite{yang2025thinking} suite of 5{,}130 questions over 288 real indoor scene videos from ScanNet, ScanNet++, and ARKitScenes, in eight categories (\cref{tab:main-results}). The four numerical categories are scored by Mean Relative Accuracy (MRA), the fraction of relative-error thresholds in $\{0.50, 0.55, \ldots, 0.95\}$ that a prediction satisfies; the four multiple-choice categories by accuracy, with the three relative-direction splits averaged into one category as in the official protocol. The overall score is the unweighted mean of the eight. We additionally evaluate on ReVSI~\cite{zhang2026revsi}, an expert-re-annotated rebuild of VSI-Bench (\cref{sec:exp-revsi}). Training and evaluation are disjoint at the scene level: none of the 288 VSI-Bench scenes occurs among the 4{,}955 training scenes, so every benchmark scene, on VSI-Bench and ReVSI alike, is unseen at every training stage. Generation is greedy with at most 16 new tokens, and answers are parsed with the official lmms-eval rules; unparseable numeric answers score 0. All models receive up to 128 frames per scene; ours additionally receives the sensor depth maps, intrinsics, and poses shipped with the benchmark scenes.

\paragraph{Reproduction gap.}
Our reproduced frame baselines fall $\sim$10 points below the official 0.5B release. The cause is our use of bf16 autocast, adopted for compute budget, in place of the official full-fp32 training. This gap only strengthens the comparison: our BEV model starts from the weaker base and still surpasses the official release by nearly 12 points. \Suppref{sec:supp-repro-gap} gives the details.

\begin{table*}
  \centering
  \caption{\textbf{VSI-Bench results} (\%). The first four categories are scored by MRA, the last four by accuracy; Avg is their unweighted mean. All models see the same $\leq$128 frames per scene.  Rows below the rule share our recipe and data (0.5B) and are trained under bf16 autocast, which accounts for their gap to the official full-fp32 releases (\suppref{sec:supp-repro-gap}); the grey row feeds BEV tokens to the frame-trained parent without fine-tuning. Bold marks the best among our rows.}
  \label{tab:main-results}
  \small
  \setlength{\tabcolsep}{3pt}
  \begin{tabular}{@{}l c r cccc cccc c@{}}
    \toprule
    Method & Encoder & \#Vis.\ tok. & \makecell{Obj.\\Count} & \makecell{Abs.\\Dist.} & \makecell{Obj.\\Size} & \makecell{Room\\Size} & \makecell{Rel.\\Dist.} & \makecell{Rel.\\Dir.} & \makecell{Route\\Plan} & \makecell{App.\\Order} & Avg \\
    \midrule
    Cambrian-S-7B~\cite{yang2026cambrians} (full fp32) & SigLIP2 & 9{,}216 & 72.81 & 49.90 & 74.72 & 69.62 & 68.45 & 74.38 & 42.27 & 81.39 & 66.69 \\
    Cambrian-S-0.5B~\cite{yang2026cambrians} (full fp32) & SigLIP2 & 9{,}216 & 68.37 & 33.51 & 53.92 & 52.92 & 55.35 & 48.98 & 29.90 & 70.87 & 51.73 \\
    \midrule
    Frames & SigLIP2 & 9{,}216 & 61.31 & 32.10 & 43.28 & 27.85 & 42.39 & 47.70 & 29.90 & 50.97 & 41.94 \\
    Frames & DINOv3 & 9{,}216 & 63.01 & 29.53 & 47.72 & 26.04 & 43.38 & 46.66 & 30.41 & 49.03 & 41.97 \\
    \textcolor{gray}{Frames parent + BEV tokens, zero-shot} & \textcolor{gray}{DINOv3} & \textcolor{gray}{729} & \textcolor{gray}{61.91} & \textcolor{gray}{32.10} & \textcolor{gray}{46.99} & \textcolor{gray}{25.90} & \textcolor{gray}{45.49} & \textcolor{gray}{46.43} & \textcolor{gray}{26.80} & \textcolor{gray}{47.25} & \textcolor{gray}{41.61} \\
    Frames + spatial FT (control) & DINOv3 & 9{,}216 & 69.79 & 37.49 & \textbf{69.97} & 60.49 & 49.86 & 48.58 & 26.80 & \textbf{76.86} & 54.98 \\
    \textbf{Cambrian-BEV (ours)} & DINOv3 & \textbf{729} & \textbf{70.16} & \textbf{44.78} & 67.36 & \textbf{85.00} & \textbf{56.20} & \textbf{82.67} & \textbf{39.18} & 62.46 & \textbf{63.48} \\
    \midrule
    $\Delta$ ours vs.\ control & -- & $12.6\times$ fewer & +0.37 & +7.29 & $-$2.61 & +24.51 & +6.34 & +34.09 & +12.38 & $-$14.40 & \textbf{+8.50} \\
    \bottomrule
  \end{tabular}
\end{table*}

\subsection{Main results}
\label{sec:exp-main}

\Cref{tab:main-results} presents the main comparison. Relative to its stage-4 parent, Cambrian-BEV gains $+21.5$ points. The frames-only control attributes $+13.0$ of this to the additional fine-tuning and $+8.5$ to the representation itself, obtained with $12.6\times$ fewer visual tokens. The zero-shot row confirms the split from the other side: BEV tokens fed to the same parent without fine-tuning score 41.6, no better than its frame tokens (42.0), so the gain requires the LLM to learn the new layout, and two epochs on VSI-288K-Depth suffice. The gains follow the geometry. Wherever the answer is a property of the metric grid, BEV wins for every configuration in \cref{tab:ablations}. Room size improves by $+24.5$, since the grid's occupied extent \emph{is} the room footprint; here our 0.5B model exceeds the official 7B ($85.0$ vs.\ $69.6$). Absolute distance ($+7.3$) and relative distance ($+6.3$) follow the same pattern. Relative direction shows the largest gain ($+34.1$), but this gain is conditional on an egocentric grid frame (\cref{sec:exp-ablations}). Route planning, which receives no training supervision (\cref{sec:exp-setup}), also improves ($+12.4$). The main cost is appearance order ($-14.4$): a time-collapsed grid cannot represent visitation order, and this category accounts for most of the remaining gap to the 7B. Prepending the frames to the grid recovers appearance order but scores 7.4 points lower overall at $13.6\times$ the tokens (\cref{tab:ablations}e, \suppref{sec:supp-ablations}).

\paragraph{Token efficiency.}
Profiled on one H100 (\suppref{sec:supp-runtime}), the $12.6\times$ shorter visual context cuts prefill $2.9\times$ ($16$ vs.\ $46$\,ms per question) and peak memory $3.6\times$ ($2.5$ vs.\ $9.1$\,GB). Building the BEV grid costs $1.9$\,ms per scene, 0.5\% of the vision encoder it already requires. The efficiency gain is therefore confined to the part of inference that depends on the visual context: prefill and memory. Decode time is unchanged, since for a 0.5B LLM it is bound by per-step launch overhead rather than context length, so the end-to-end saving per question is $1.13\times$.

\subsection{Ablations}
\label{sec:exp-ablations}

\begin{table}
  \centering
  \caption{\textbf{Ablations}. Single-variable changes to the recipe of \cref{tab:main-results}, on VSI-Bench (overall average, room size, relative direction) and ReVSI (overall average, \emph{all\_frame}, \cref{sec:exp-revsi}). ``FP'' is first-pose orientation. Bold marks the best in each column. We adopt hard$\times$FP as the default because it ties bilinear$\times$FP on VSI-Bench with a $3.4\times$ cheaper projection (\cref{sec:exp-ablations}); bilinear$\times$FP is the stronger variant on ReVSI. Blocks (d) and (e) are discussed in \suppref{sec:supp-ablations}.}
  \label{tab:ablations}
  \small
  \setlength{\tabcolsep}{2pt}
  \begin{tabular}{@{}l c c c c@{}}
    \toprule
    & \multicolumn{3}{c}{VSI-Bench} & ReVSI \\
    \cmidrule(lr){2-4} \cmidrule(lr){5-5}
    Variant & Avg & Room & Rel.\ Dir. & Avg \\
    \midrule
    \multicolumn{5}{@{}l}{\emph{(a) Grid frame: aggregation $\times$ orientation}} \\
    hard $\times$ world & 55.69 & 84.44 & 46.87 & 41.79 \\
    hard $\times$ PCA-rectified & 52.09 & 82.92 & 47.52 & 41.16 \\
    \textbf{hard $\times$ FP (ours)} & \textbf{63.48} & 85.00 & \textbf{82.67} & 44.68 \\
    bilinear $\times$ world & 60.32 & 81.81 & 78.27 & 45.20 \\
    bilinear $\times$ FP & 63.39 & 84.79 & 77.49 & \textbf{46.39} \\
    \midrule
    \multicolumn{5}{@{}l}{\emph{(b) Encoder (hard $\times$ FP)}} \\
    DINOv3 ViT-L/16 & \textbf{63.48} & 85.00 & \textbf{82.67} & 44.68 \\
    SigLIP2-So400M & 55.06 & 84.41 & 45.64 & 40.56 \\
    \midrule
    \multicolumn{5}{@{}l}{\emph{(c) Grid resolution at $\pm5$\,m (hard $\times$ FP)}} \\
    $18{\times}18$ (324 tok.) & 55.70 & 83.85 & 48.48 & 41.02 \\
    $27{\times}27$ (729 tok.) & \textbf{63.48} & 85.00 & \textbf{82.67} & 44.68 \\
    $36{\times}36$ (1{,}296 tok.) & 57.90 & \textbf{85.90} & 49.09 & 43.45 \\
    \midrule
    \multicolumn{5}{@{}l}{\emph{(d) Learnable interface components (hard $\times$ FP)}} \\
    + 2-layer MLP adapter & 51.45 & 84.17 & 46.34 & 41.11 \\
    + grid self-attention encoder & 44.69 & 70.52 & 46.97 & 35.47 \\
    + learned empty-cell token & 60.07 & 84.93 & 77.00 & 43.38 \\
    \midrule
    \multicolumn{5}{@{}l}{\emph{(e) Alternative representations (hard $\times$ FP)}} \\
    16 temporal chunks + global grid & 54.44 & 84.69 & 45.17 & 40.48 \\
    hybrid: 128 frames + BEV & 56.06 & 82.50 & 47.18 & 40.78 \\
    height filter 0.05--2.0\,m & 57.55 & 81.53 & 50.99 & 42.51 \\
    \bottomrule
  \end{tabular}
\end{table}

\Cref{tab:ablations} collects the single-variable ablations, organized around three questions: how the grid should be defined (a, c), what should be pooled into it (b), and whether the parameter-free interface benefits from added capacity or alternative representations (d, e). Blocks (d) and (e) are discussed in \suppref{sec:supp-ablations}; none of their variants improves on the plain grid.

\paragraph{(a) Grid frame: orientation and aggregation.}
Room size is nearly flat across the table (81.5--85.9 for all but one BEV row, against 60.5 for the frames-only control): the grid's metric content depends little on how it is oriented, splatted, or sized. Relative direction behaves in the opposite way. It requires a consistent egocentric reference frame, which the two grid-frame levers supply by different routes. \emph{Orientation}: rotating the grid so that its forward axis follows the first camera's heading lifts the overall score from 55.7 to 63.5 and relative direction from $46.9$ to $82.7$, the largest single effect in the paper, because a trajectory-anchored frame gives directional language (``to the left of'') a consistent geometric meaning across training scenes, whereas world or scene-PCA axes (\suppref{sec:supp-projection}) make it arbitrary per scene. \emph{Aggregation}: with world axes fixed, bilinear splatting alone recovers most of the same effect (relative direction $78.3$); we report this without isolating the mechanism. The two levers are not additive on VSI-Bench (63.5 vs.\ 63.4): relative direction saturates once \emph{either} defect is fixed. We adopt hard$\times$first-pose as the default because it ties on VSI-Bench with a $3.4\times$ cheaper projection (\suppref{sec:supp-runtime}) and the simpler operator; bilinear splatting is less sensitive to the grid frame and scores higher on ReVSI in every configuration (\cref{sec:exp-revsi}), and we recommend it when projection cost is not a constraint.

\paragraph{(b) Encoder.}
At matched geometry, replacing DINOv3 with SigLIP2 costs 8.4 points on VSI-Bench and 4.1 on ReVSI, concentrated in relative direction ($82.7$ vs.\ $45.6$). Since the two encoders tie as frame-token models (42.0 vs.\ 41.9, \cref{tab:main-results}), this is the end-to-end signature of the feature-locality argument of \cref{sec:method-encoder}: both encoders benefit from the BEV representation, but the locally grounded one benefits dramatically more.

\paragraph{(c) Grid resolution has a narrow optimum on VSI-Bench.}
At a fixed $\pm5$\,m range, $27{\times}27$ (0.37\,m cells) is best. Both $18{\times}18$ and $36{\times}36$ lose the directional capability. The optimum is less pronounced on ReVSI, where the same three checkpoints span 3.7 points against 7.8 on VSI-Bench, so $27{\times}27$ is a well-supported default rather than a fragile requirement.

\subsection{Results on ReVSI}
\label{sec:exp-revsi}

\begin{table}
  \centering
  \caption{\textbf{ReVSI results}. Avg is the unweighted mean of the seven categories; both subsets cover the same 380 scenes. \emph{all\_frame}: 6{,}808 questions, 128 frames per scene; \emph{64\_frame}: 6{,}616 questions, 64 frames per scene. Per-category scores are in \suppref{sec:supp-revsi}. All rows are 0.5B; ours are trained under bf16 autocast (\suppref{sec:supp-repro-gap}). Bold marks the best among our rows.}
  \label{tab:revsi-results}
  \small
  \begin{tabular}{@{}lcc@{}}
    \toprule
    Method & \emph{all\_frame} & \emph{64\_frame} \\
    \midrule
    Cambrian-S-0.5B (official, full fp32) & 37.68 & 37.59 \\
    \midrule
    Frames (SigLIP2) & 32.10 & 31.92 \\
    Frames (DINOv3) & 33.41 & 33.10 \\
    Frames + spatial FT (control) & 40.17 & 40.02 \\
    BEV, SigLIP2 (hard $\times$ FP) & 40.56 & 40.34 \\
    BEV, hard $\times$ world & 41.79 & 42.27 \\
    \textbf{Cambrian-BEV (hard $\times$ FP)} & 44.68 & 45.00 \\
    BEV, bilinear $\times$ world & 45.20 & 45.02 \\
    BEV, bilinear $\times$ FP & \textbf{46.39} & \textbf{46.21} \\
    \bottomrule
  \end{tabular}
\end{table}

VSI-Bench inherits noisy 3D annotations from its source reconstructions. ReVSI~\cite{zhang2026revsi} is an expert re-annotated rebuild that adds 3RScan and MultiScan captures to VSI-Bench's three sources. \Cref{tab:revsi-results} evaluates the same checkpoints, without retraining, under both its 128-frame and 64-frame protocols.

The main conclusions transfer. Every BEV variant beats the frames-only control; our default leads it by $+4.5$ (44.7 vs.\ 40.2), room size is again the diagnostic category (58.7 vs.\ 45.3), the encoder-locality gap replicates ($+4.1$), and BEV again wins route planning (49.5 vs.\ 42.1) without route supervision. The gains are also more uniform across recipes: relative direction varies by 11 points across our grid configurations on ReVSI against 36 on VSI-Bench, so on the expert-annotated benchmark the BEV advantage does not hinge on a single geometry choice. Bilinear splatting with first-pose orientation adds $+1.7$ over our default and leads six of the seven categories, and the 64-frame budget reproduces the same ordering (\cref{tab:revsi-results}).

\subsection{Qualitative analysis}
\label{sec:exp-qualitative}

\begin{figure}
  \centering
  \includegraphics[width=\linewidth]{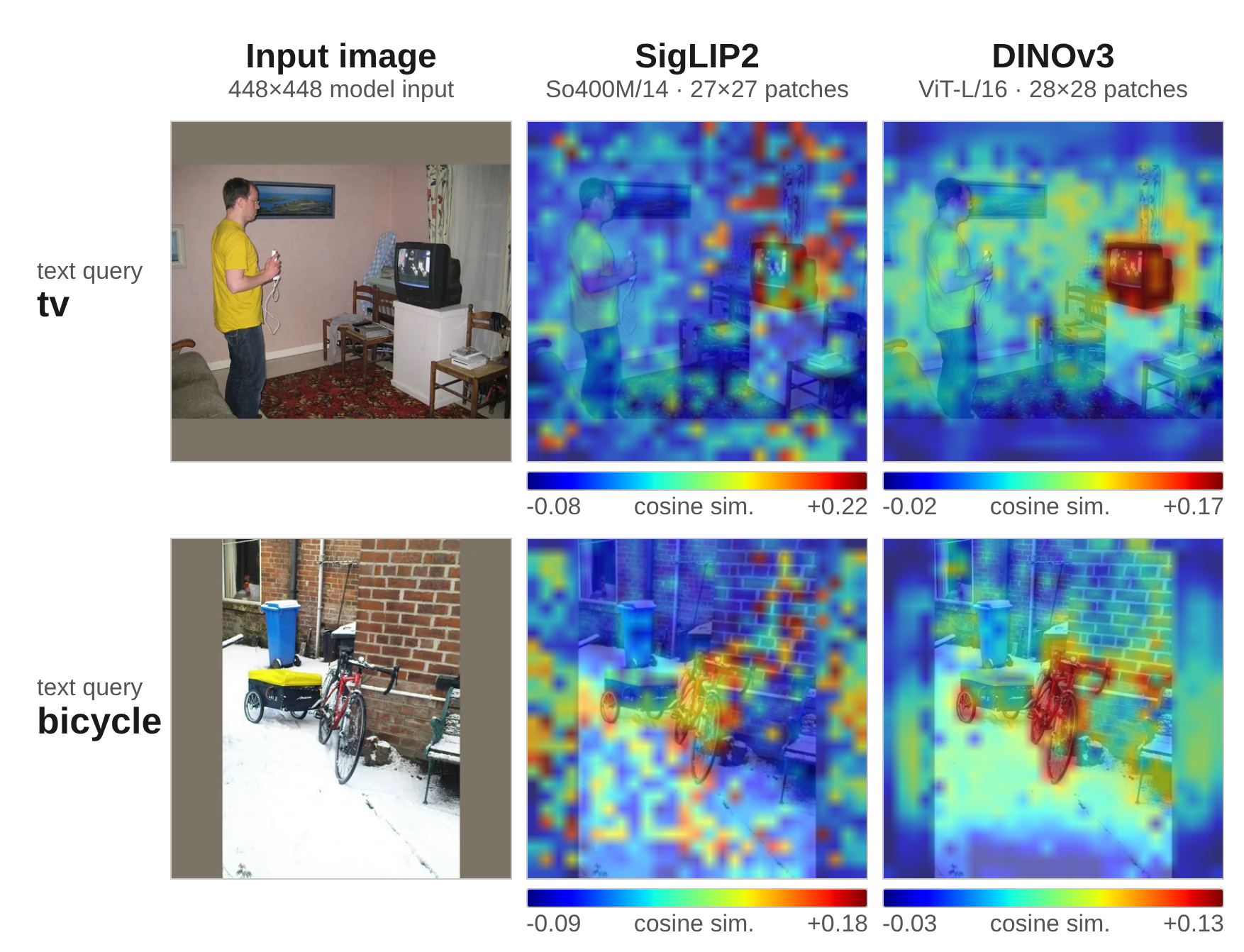}
  \caption{\textbf{Patch-level text similarity, SigLIP2 vs.\ DINOv3.} SigLIP2 distributes evidence for the queried object (row label) across the image, whereas DINOv3 more closely follows the object's extent. Warm colors indicate higher similarity; color scales are normalized per panel. Details in \suppref{sec:supp-text-sim}.}
  \label{fig:text-sim}
\end{figure}

\paragraph{Feature locality.}
\Cref{fig:text-sim} compares patch-level text-similarity maps from SigLIP2 and DINOv3 (probe details in \suppref{sec:supp-text-sim}). SigLIP2 distributes high similarity across object and background regions, whereas DINOv3 more closely follows the queried object's extent. This is the locality that BEV aggregation requires (\cref{sec:method-encoder}) and is the mechanism behind the 8.4-point encoder gap in \cref{tab:ablations}. Frame-token accuracy therefore cannot be used to select an encoder for geometric fusion.

\paragraph{Qualitative examples and failure modes.}
\Suppref{sec:supp-qualitative} shows BEV occupancy on three VSI-Bench scenes with room-size and relative-direction questions that the frames-only control answers incorrectly and Cambrian-BEV answers correctly (\suppref{fig:supp-qualitative}), and the three characteristic failure modes of the representation: lost visitation order, sub-cell objects, and corrupted geometry (\suppref{fig:supp-failures}).

\section{Conclusion}
\label{sec:conclusion}

We introduced Cambrian-BEV, a parameter-free geometric tokenization that uses depth and camera pose to aggregate video features into a shared BEV grid. By replacing $9{,}216$ frame tokens with 729 BEV tokens, our method shrinks the LLM visual context $12.6\times$, cutting prefill time $2.9\times$ and peak memory $3.6\times$, while improving spatial reasoning on VSI-Bench and ReVSI. We further showed that frame-token performance does not predict performance under geometric rearrangement: two encoders that tie as frame-token models diverge substantially once their patches are fused in BEV, making dense feature locality a necessary selection criterion for geometry-aware MLLMs. The representation has clear limits: a single time-collapsed grid discards visitation order, the one category where frame tokens remain clearly ahead, and flattens vertical structure, so thin or stacked objects share a cell with the surfaces behind them; simple remedies such as temporal chunking, frame--BEV hybrids, and height filtering did not close these gaps (\suppref{sec:supp-ablations}). The method also assumes calibrated depth and pose and is best suited to static, room-scale indoor scenes. Future work will pursue temporal encodings that preserve grid resolution, height-aware pooling, robustness to estimated geometry, and larger or dynamically expanding grids for longer trajectories and larger backbones.

{
    \small
    \bibliographystyle{ieeenat_fullname}
    \bibliography{main}
}

\clearpage
\appendix
\setcounter{figure}{0}\renewcommand{\thefigure}{S\arabic{figure}}
\setcounter{table}{0}\renewcommand{\thetable}{S\arabic{table}}
\setcounter{equation}{0}\renewcommand{\theequation}{S\arabic{equation}}
\maketitlesupplementary

\section{BEV projection details}
\label{sec:supp-projection}
This section gives a complete specification of the parameter-free BEV module of \cref{sec:method-projection}, including the coordinate conventions, default values, and degenerate cases that the main text only summarizes. All choices below are those of the final model unless stated otherwise; the alternatives are the ablation arms of \cref{tab:ablations}.

\paragraph{Inputs and rescaling.}
A scene contributes $T{=}128$ frames sampled uniformly over the recorded trajectory (\texttt{linspace} over the frame indices, so frame 0 of the video is always the first sampled frame; scenes shorter than 128 frames contribute all of their frames). Each frame provides an RGB image, a 16-bit depth map in millimeters, intrinsics $K_t$, and a world-to-camera pose $T^{w2c}_t$. RGB and depth are resized to the encoder resolution of $448{\times}448$ without preserving aspect ratio (bicubic for RGB, nearest-neighbor for depth so that no interpolated depth values are created), and the intrinsics are rescaled to match: the first row of $K_t$ is multiplied by $448/W_{\mathrm{native}}$ and the second row by $448/H_{\mathrm{native}}$, so the anisotropic resize introduces no geometric error. The encoder (DINOv3 ViT-L/16) yields a $28{\times}28$ patch grid per frame, so a scene produces $128{\times}784 = 100{,}352$ candidate points, each carrying one projector-output token $f_p \in \mathbb{R}^{896}$.

\paragraph{Depth sampling and validity.}
The depth of patch $(i, j)$ is read at the patch center $(u, v) = ((j{+}\tfrac12)P, (i{+}\tfrac12)P)$, $P{=}16$, by bilinear interpolation of the resized depth map (border padding). A patch is discarded if its sampled depth is non-finite, non-positive, or outside $[0.1, 100]$\,m; missing sensor depth is stored as zero and therefore falls in this category. A patch whose center lands in a depth hole is dropped even when its RGB content is valid. We also implemented a masked average-pooling variant that averages all valid depth pixels inside the $16{\times}16$ patch footprint; it is more robust to speckle but was not used in any reported run.

\paragraph{Common world frame.}
The three capture sources ship poses in different conventions. We convert all of them to a single right-handed, gravity-aligned world frame with $Z$ up and an OpenCV camera ($X$ right, $Y$ down, $Z$ forward). ScanNet and ScanNet++ scan frames are already $Z$-up; ARKitScenes poses are expressed with an OpenGL camera ($Y$ up, $Z$ backward) in a $Y$-up world and are converted with $\mathrm{diag}(1,-1,-1)$ on the camera side and a $90^\circ$ rotation about $X$ on the world side (the shipped \texttt{.align.json} transforms do not give a clean gravity-aligned frame and are ignored). The vertical axis is therefore consistent across datasets, which is what the height statistics of \cref{sec:method-projection} and the floor anchor of \cref{sec:supp-height-filter} rely on, but the horizontal axes and the origin remain whatever each capture pipeline produced: ScanNet frames are roughly aligned to the room's walls, ScanNet++ and ARKitScenes frames are not, and the origin may lie anywhere relative to the room. In the final model, nothing downstream depends on the absolute horizontal position or heading of the world frame; both are removed by the centering and orientation steps below. The \emph{world} orientation arm of \cref{tab:ablations} is the only configuration in which the raw horizontal axes reach the grid.

\paragraph{Orientation.}
Only the horizontal coordinates $(x, y)$ of the world points are used for binning; height is dropped after the optional filter. Before centering, they are rotated by a per-scene $2{\times}2$ rotation $R_\theta$, where $\theta$ is chosen by one of three rules. \emph{World}: $R_\theta = I$. \emph{First-pose}: the first camera's optical axis is carried to the world frame, $\mathbf{d} = R_0^\top (0, 0, 1)^\top$, projected onto the ground plane, and normalized, $\hat{\mathbf{d}} = (d_x, d_y)/\lVert (d_x, d_y) \rVert$; the rotation maps $\hat{\mathbf{d}}$ to the grid's forward axis,
\begin{equation}
  R_\theta = \begin{bmatrix} \hat d_x & \hat d_y \\ -\hat d_y & \hat d_x \end{bmatrix}, \qquad R_\theta \hat{\mathbf{d}} = (1, 0)^\top.
  \label{eq:first-pose-rot}
\end{equation}
Only the yaw of the first camera enters; its pitch and roll are discarded by the projection, and its position is not used (see the centering paragraph). Because the first sampled frame is frame 0 of the video, the grid frame is a deterministic function of the scene and does not depend on how many frames are sampled. \emph{PCA-rectified}: let $\bar{\mathbf{x}}$ be the mean of the valid horizontal points; points farther than the 95th percentile of distances from $\bar{\mathbf{x}}$ are trimmed, the $2{\times}2$ covariance of the remaining points is diagonalized, and its leading eigenvector $\mathbf{e}_1$ takes the place of $\hat{\mathbf{d}}$ in \cref{eq:first-pose-rot}. Eigenvectors are defined up to sign, so $\mathbf{e}_1$ is flipped to have a non-negative dot product with the mean over all frames of the cameras' ground-plane forward directions; scenes with fewer than three valid points fall back to $R_\theta = I$. PCA thus aligns the grid with the room's long axis, a property of the scene geometry, whereas first-pose aligns it with the observer, a property of the trajectory; \cref{tab:ablations} shows that only the latter gives relative-direction language a consistent meaning. In every mode $R_\theta$ is a proper rotation about the vertical axis, so left and right are never mirrored.

\paragraph{Centering.}
After rotation, the grid is centered on the centroid $\mathbf{c} = \tfrac{1}{|\mathcal{V}|}\sum_{p \in \mathcal{V}} (x_p, y_p)$ of the valid points $\mathcal{V}$, that is, of all patches that survived the depth test and the optional height filter, weighted equally per patch. The centroid is computed \emph{before} the $\pm5$\,m crop, so a scene is centered on everything it observed, including distant outliers; this is why corrupted poses can move the grid away from the room entirely (\cref{fig:supp-failures}c). It is also a feature-mass centroid rather than a geometric one: surfaces observed in many frames pull the center toward themselves. Rotation and centering commute, so the grid frame is equivalently ``rotate about the centroid.'' The grid is \emph{not} centered on the first camera: in the final model the orientation is egocentric (first-pose yaw) but the position is scene-centric (centroid).

\paragraph{Binning and token layout.}
With $\mathbf{x}' = R_\theta \mathbf{x} - \mathbf{c}$ the rotated, centered coordinates, a point falls in row $r = \lfloor (5 - x')/\Delta \rfloor$ and column $k = \lfloor (5 - y')/\Delta \rfloor$ with $\Delta = 10/27 = 0.370$\,m. Row 0 is therefore the most forward strip of the scene and column 0 the leftmost, so the grid, read as a matrix, is a top-down map drawn with forward at the top and the observer's left on the left. Points with $r$ or $k$ outside $[0, 26]$ are dropped (1.1\% of points, \cref{tab:supp-grid-range}). The $27{\times}27$ cells are flattened row-major into $729$ tokens, token $n$ being cell $(\lfloor n/27 \rfloor, n \bmod 27)$, and spliced into the sequence in place of the \texttt{<image>} placeholder with no separator, newline token, or positional embedding of their own; the only positional signal the LLM receives is its own rotary embedding over the sequence index, from which it must infer the $27$-wide row structure.

\paragraph{Hard assignment.}
Each surviving point deposits its token in the single cell that contains it. All $B$ scenes of a batch are scattered in one \texttt{scatter\_add} into a flat $[B{\cdot}729, 896]$ buffer, with a parallel scatter of ones giving the per-cell count $n_{rk}$. The cell token is $B_{rk} = \tfrac{1}{n_{rk}} \sum_{p \in \mathcal{P}_{rk}} f_p$ (\cref{eq:aggregation}).

\paragraph{Bilinear splatting.}
In the alternative aggregation, the continuous grid coordinate $(\rho, \kappa) = ((5 - x')/\Delta, (5 - y')/\Delta)$ is split into an integer part $(r_0, k_0) = (\lfloor\rho\rfloor, \lfloor\kappa\rfloor)$ and fractional part $(\delta_r, \delta_k)$, and the token is distributed to the four cells $(r_0{+}a, k_0{+}b)$, $a, b \in \{0, 1\}$, with weights $w_{ab} = \delta_r^{a}(1{-}\delta_r)^{1-a}\,\delta_k^{b}(1{-}\delta_k)^{1-b}$. Weights are accumulated alongside the features and each cell is normalized by its accumulated weight, so the result is again a weighted mean. Corner cells that fall outside the grid are dropped individually, so a point in the border strip loses part of its mass. The weights are defined with respect to the integer lattice of cell origins, so a point on the boundary between two hard-assignment cells is split evenly between them, whereas a point exactly on a cell origin contributes to that cell only. Splatting touches four cells per point and runs four scatters, which is the source of the $6\times$ cost difference in the scatter ($5.3$ vs.\ $0.9$\,ms per scene on an H100; $6.4$ vs.\ $1.9$\,ms for the full projection).

\paragraph{Empty cells and occupancy.}
\phantomsection\label{sec:supp-empty-cells}
Cells that receive no point keep a zero vector and are recorded in a Boolean occupancy mask. In the final model the zero tokens are passed to the LLM unchanged and are attended like any other token; the count of empty cells is itself informative (it is the complement of the room footprint). The ablations of \cref{tab:ablations} replace the zero vector with a learned token, which does not improve on the plain grid.

\paragraph{Numerics and gradients.}
All geometry (depth sampling, unprojection, pose transform, rotation, centering, binning) runs in fp32 with autocast disabled and under \texttt{no\_grad}; it produces only indices, weights, and masks. The scatter itself is applied to the tokens in their training dtype (bf16), while the per-cell counts and splat weights are accumulated in fp32, because a bf16 counter cannot exceed $256$: above that value the representable numbers are two apart, so adding one rounds back to $256$. Because scatter-add followed by count normalization is linear in the tokens, the gradient that reaches a patch token is the gradient of its cell divided by the cell count (or scaled by its splat weight). The module therefore has no parameters but is fully differentiable with respect to the projector, which is trained through it.

\paragraph{Degenerate cases.}
(i) A first camera pointing straight up or down has a vanishing ground-plane heading; the normalization in \cref{eq:first-pose-rot} is guarded with a small $\epsilon$, but the resulting yaw is arbitrary; handheld and head-mounted recordings start roughly level, and we apply no further guard. (ii) Pose drift or gross pose errors make the centroid meaningless and can push every point outside the $\pm5$\,m window, leaving all 729 tokens empty (\cref{fig:supp-failures}c); the model then answers from its prior. (iii) Scenes in which a large fraction of patches have no depth (reflective or distant surfaces) lose those patches entirely rather than receiving an approximate location.

\section{Height-filter details}
\label{sec:supp-height-filter}
The floor-relative height filter of \cref{sec:method-projection} estimates a per-scene floor anchor as the 2nd percentile of valid world-frame point heights and drops points outside $[z_{\min}, z_{\max}]$ above it. A per-scene anchor is necessary because absolute height references differ across capture sources: ScanNet-family scenes are floor-referenced whereas ARKitScenes scenes are device-referenced, so no global threshold works. The band evaluated in \cref{tab:ablations} is $[0.05, 2.0]$\,m, which removes floor speckle and ceiling mass; it costs 5.9 points on the VSI-Bench average and is therefore disabled in the final model.

\section{Ablations (d) and (e): added capacity and alternative representations}
\label{sec:supp-ablations}
This section discusses blocks (d) and (e) of \cref{tab:ablations}, which the main text only summarizes. All variants are single-variable changes to the final recipe (hard assignment, first-pose orientation, $27{\times}27$ grid at $\pm5$\,m) and are compared against its 63.5 VSI-Bench average.

\paragraph{(d) Added parameters do not help.}
Every learnable component we tried costs performance relative to the parameter-free interface, all measured on the first-pose grid against the 63.5 anchor: a 2-layer MLP adapter between pooled tokens and the LLM ($-12.0$), a zero-initialized axial-RoPE self-attention encoder over the grid with trainable newline tokens ($-18.8$), and a learned token for empty cells ($-3.4$). The first two also erase the orientation effect: relative direction falls from $82.7$ to $46.3$ and $47.0$, the world-frame level of (a), so the modules are not merely non-additive with first-pose but mask it. The encoder is the only variant that also degrades the metric categories (room size $85.0 \to 70.5$), and it is worst on ReVSI as well ($-9.2$). Pooled projector outputs are already language-space vectors, and randomly initialized modules only disturb them under the short fine-tuning budget.

\paragraph{(e) Alternative representations.}
Three natural modifications all underperform the single grid. \emph{Temporal chunking}: 16 per-chunk grids plus a global grid leave relative direction at world level ($45.2$); the first-pose benefit does not transfer to chunked grids. \emph{Hybrid frames{+}BEV}: prepending up to 128 frames to the first-pose grid scores 56.1, $7.4$ points below BEV-only at $13.6\times$ the token budget, with relative direction back at world level ($47.2$ vs.\ $82.7$). The only category the hybrid wins is appearance order, which is exactly what frames in temporal order supply. \emph{Height filtering}: a floor-relative 0.05--2.0\,m band costs $-5.9$ and drops relative direction from $82.7$ to $51.0$, even though the vertical-dilution analysis of \cref{sec:method-projection} predicts a gain. We hypothesize, without testing it here, that the floor and ceiling mass keeps the full scene footprint represented in the grid, and that this geometric context outweighs the dilution it causes.

\section{Grid range}
\label{sec:supp-grid-range}
The $\pm5$\,m range of the BEV grid is a data-driven choice. For each of the 4{,}955 scenes of VSI-288K-Depth we back-projected 128 uniformly sampled frames with the final model's geometry (DINOv3 at $448{\times}448$, patch 16, first-pose orientation), centered the points on their centroid as in \cref{sec:method-projection}, and measured the fraction of points falling inside a symmetric $\pm R$ window for $R \in \{3, 4, 5, 6, 7.5, 10\}$\,m (\cref{tab:supp-grid-range}). At $\pm5$\,m the grid retains 98.9\% of all observed points, with a median per-scene retention above 99.8\% on every source dataset, and 96\% of scenes retain at least 95\% of their points. Shrinking the window discards scene content quickly: $\pm4$\,m drops 2.9\% of points and $\pm3$\,m 8.7\%, with ScanNet++ scenes losing 13\%. Widening it buys little: $\pm7.5$\,m recovers a further 0.9\% of points, but at a fixed $27{\times}27$ resolution it enlarges each cell from 0.37 to 0.56\,m, so the same tokens carry a coarser description of the scene.

\begin{table}[h]
  \centering
  \small
  \setlength{\tabcolsep}{3pt}
  \caption{\textbf{Grid range vs.\ coverage.} Fraction (\%) of valid back-projected points inside a centered $\pm R$ window over the 4{,}955 scenes of VSI-288K-Depth, and cell size at $27{\times}27$. Bold marks the range used by the final model.}
  \label{tab:supp-grid-range}
  \begin{tabular}{@{}lccccc@{}}
    \toprule
    Range & Cell (m) & ScanNet & ScanNet++ & ARKitScenes & All \\
    \midrule
    $\pm3$\,m   & 0.22 & 93.79 & 86.98 & 91.75 & 91.29 \\
    $\pm4$\,m   & 0.30 & 98.43 & 94.60 & 97.47 & 97.14 \\
    \textbf{\boldmath$\pm5$\,m} & \textbf{0.37} & \textbf{99.55} & \textbf{97.40} & \textbf{99.19} & \textbf{98.94} \\
    $\pm6$\,m   & 0.44 & 99.85 & 98.59 & 99.72 & 99.54 \\
    $\pm7.5$\,m & 0.56 & 99.97 & 99.30 & 99.94 & 99.83 \\
    $\pm10$\,m  & 0.74 & 100.00 & 99.61 & 99.99 & 99.92 \\
    \bottomrule
  \end{tabular}
\end{table}

\section{Inference cost}
\label{sec:supp-runtime}
\Cref{tab:runtime} profiles where inference time goes for Cambrian-BEV and the frames-only control, measured on one H100 (SXM) with CUDA events, bf16, FlashAttention-3 and a fixed 15-step decode budget (medians over 96 questions on 8 VSI-Bench scenes with exactly 128 frames each; the two arms differ only in the visual tokens they present to the LLM). The cost splits into a \emph{per-scene} part, paid once and cached across a scene's questions, and a \emph{per-question} part. Per scene, the DINOv3 encoder and projector dominate ($365$\,ms, 99\% of scene encoding) and is identical for both models; constructing the BEV grid from $128{\times}784$ patch features (depth sampling, unprojection, grid assignment and scatter) adds $1.9$\,ms, 0.5\% of scene encoding and within the run-to-run noise of the encoder. The bilinear-splatting variant of \cref{tab:ablations} raises this to $6.4$\,ms (the scatter alone is $6.1\times$ slower, $5.3$ vs.\ $0.9$\,ms), still 1.7\% of scene encoding. Per question, the $12.6\times$ shorter visual context cuts LLM prefill by $2.9\times$ ($15.7$ vs.\ $45.9$\,ms; the $+30$\,ms gap is stable to $\pm 0.3$\,ms across two independent runs) and peak memory by $3.6\times$ ($2.5$ vs.\ $9.1$\,GB). The saving is sub-linear in tokens because at these lengths prefill is not yet attention-dominated. Decode is unchanged ($211$\,ms for 15 steps for both models, with run-to-run drift larger than the model-to-model difference), since for a 0.5B LLM it is bound by per-step launch overhead rather than by context length, and it dilutes the end-to-end gap to $1.13\times$ ($236$ vs.\ $267$\,ms per question). Because small launch-bound kernels are the most hardware-sensitive, the ratios depend on the accelerator: on an H200 NVL the same code gives a $2.1\times$ prefill saving, so ratios should be quoted with the hardware. The representation therefore costs nothing to build on top of the encoder it already requires and removes 66\% of prefill time and 72\% of peak memory; further per-question savings would have to come from the decode loop, not from the visual representation.

\begin{table}
  \centering
  \caption{\textbf{Inference cost} of Cambrian-BEV and the frames-only control, median GPU ms on one H100 SXM (bf16, FlashAttention-3, 128-frame scenes, 15 decode steps). Scene encoding is paid once per scene and cached across its questions; the remaining rows are paid per question. Peak memory is measured over the whole pipeline.}
  \label{tab:runtime}
  \small
  \begin{tabular*}{\columnwidth}{@{\extracolsep{\fill}} l r r r @{}}
    \toprule
    & Frames & \textbf{BEV} & Ratio \\
    \midrule
    Visual tokens & 9{,}216 & \textbf{729} & $12.6\times$ \\
    \midrule
    \multicolumn{4}{@{}l}{\emph{Per scene} (ms)} \\
    \quad Vision encoder + projector & 366.0 & 364.8 & $1.0\times$ \\
    \quad Frame downsample / BEV build & 0.4 & 1.9 & -- \\
    \midrule
    \multicolumn{4}{@{}l}{\emph{Per question} (ms)} \\
    \quad LLM prefill & 45.9 & \textbf{15.7} & $\mathbf{2.9\times}$ \\
    \quad LLM decode (15 steps) & 211.4 & 211.0 & $1.0\times$ \\
    \quad End-to-end & 267.0 & 236.2 & $1.13\times$ \\
    \midrule
    Peak GPU memory (GB) & 9.1 & \textbf{2.5} & $\mathbf{3.6\times}$ \\
    \bottomrule
  \end{tabular*}
\end{table}

\section{Training recipe}
\label{sec:supp-stages}
\Cref{tab:supp-stages} lists the four Cambrian-S stages as run in our reproduction, and the fifth stage that produces Cambrian-BEV and the frames-only control. Data at stages 1--4 are the complete released sets~\cite{yang2026cambrians}; the vision encoder is frozen at every stage, and all stages use a cosine schedule under bf16 autocast (\cref{sec:supp-repro-gap}). Stage~5 differs from stage~4 only in restricting the data to VSI-288K-Depth (\cref{sec:exp-setup}) and, for Cambrian-BEV, in replacing the frame tokens with BEV tokens.

\begin{table*}[t]
  \centering
  \small
  \setlength{\tabcolsep}{4pt}
  \begin{tabular}{@{}llllrrr@{}}
    \toprule
    Stage & Data & Trainable & Frames & Seq.\ len. & LR & Ep. \\
    \midrule
    1 Alignment  & Cambrian-Alignment (2.5M)                                   & Proj.\       & 1   & 2{,}048  & $10^{-3}$ & 1 \\
    2 Image IT   & Cambrian-7M                                                 & Proj.\ + LLM & 1   & 8{,}192  & $10^{-5}$ & 1 \\
    3 Video IT   & Cambrian-S-3M + Cambrian-7M (737K)                            & Proj.\ + LLM & 64  & 8{,}192  & $10^{-5}$ & 1 \\
    4 Spatial IT & VSI-590K + Cambrian-S-3M (590K) + Cambrian-7M (120K) & Proj.\ + LLM & 128 & 16{,}384 & $10^{-5}$ & 1 \\
    \midrule
    5 BEV / frames FT & VSI-288K-Depth                                         & Proj.\ + LLM & 128 & 16{,}384 & $2{\times}10^{-5}$ & 2 \\
    \bottomrule
  \end{tabular}
  \caption{\textbf{Training recipe.} Stages 1--4 are the Cambrian-S recipe as run in our reproduction; stage~5 is the fine-tuning stage of \cref{sec:exp-setup}, run once with frame tokens (control) and once with BEV tokens (Cambrian-BEV). The vision encoder is frozen throughout. Parenthesized sizes are subsets of the named corpus.}
  \label{tab:supp-stages}
\end{table*}

\begin{table}[t]
  \centering
  \small
  \caption{Statistics of each panel of \cref{fig:text-sim}: cosine similarity over all patch tokens, with $p_{2}$ and $p_{98}$ the endpoints of that panel's colorbar, and Moran's $I$ (4-neighbour spatial autocorrelation). Grids are $27{\times}27$ (SigLIP2) and $28{\times}28$ (DINOv3).}
  \label{tab:supp-text-sim}
  \setlength{\tabcolsep}{2pt}
  \begin{tabular*}{\columnwidth}{@{\extracolsep{\fill}} llrrrrrr @{}}
    \toprule
    Query & Encoder & min & mean & max & $p_{2}$ & $p_{98}$ & $I$ \\
    \midrule
    ``tv'' & SigLIP2 & $-0.105$ & $-0.001$ & $0.255$ & $-0.084$ & $0.216$ & $0.33$ \\
    ``tv'' & DINOv3 & $-0.058$ & $0.041$ & $0.214$ & $-0.021$ & $0.167$ & $0.75$ \\
    ``bicycle'' & SigLIP2 & $-0.167$ & $0.001$ & $0.219$ & $-0.088$ & $0.176$ & $0.33$ \\
    ``bicycle'' & DINOv3 & $-0.058$ & $0.024$ & $0.200$ & $-0.033$ & $0.125$ & $0.74$ \\
    \bottomrule
  \end{tabular*}
\end{table}

\begin{table*}[t]
  \centering
  \caption{\textbf{ReVSI \emph{all\_frame} per category} (6{,}808 questions, 380 scenes, 128 frames per scene). Avg is the unweighted mean of the seven categories. All rows are 0.5B. Bold marks the best among our rows.}
  \label{tab:supp-revsi-allframe}
  \small
  \begin{tabular*}{\textwidth}{@{\extracolsep{\fill}} l rrrrrrr r @{}}
    \toprule
    Method & Obj.\ Count & Abs.\ Dist. & Obj.\ Size & Room Size & Rel.\ Dist. & Rel.\ Dir. & Route & Avg \\
    \midrule
    Cambrian-S-0.5B (official, full fp32) & 40.77 & 32.92 & 46.30 & 37.72 & 34.20 & 35.54 & 36.33 & 37.68 \\
    \midrule
    Frames (SigLIP2) & 32.14 & 38.36 & 32.22 & 25.36 & 30.05 & 34.75 & 31.83 & 32.10 \\
    Frames (DINOv3) & 32.44 & 36.90 & 34.98 & 31.39 & 30.35 & 36.63 & 31.19 & 33.41 \\
    Frames + spatial FT (control) & 38.91 & 38.12 & 47.34 & 45.30 & 33.01 & 36.41 & 42.12 & 40.17 \\
    BEV, SigLIP2 (hard $\times$ FP) & 37.34 & 36.00 & 46.83 & 54.37 & 31.31 & 36.27 & 41.80 & 40.56 \\
    BEV, hard $\times$ world & \textbf{39.44} & 39.16 & 46.44 & 58.54 & 31.53 & 34.68 & 42.77 & 41.79 \\
    \textbf{Cambrian-BEV (hard $\times$ FP)} & 38.09 & 41.14 & 46.80 & 58.74 & 34.72 & 43.76 & 49.52 & 44.68 \\
    BEV, bilinear $\times$ world & 38.52 & 40.34 & \textbf{47.40} & 59.50 & 35.83 & 45.64 & 49.20 & 45.20 \\
    BEV, bilinear $\times$ FP & 38.50 & \textbf{44.01} & 47.37 & \textbf{61.19} & \textbf{37.45} & \textbf{46.07} & \textbf{50.16} & \textbf{46.39} \\
    \bottomrule
  \end{tabular*}
\end{table*}
\begin{table*}[t]
  \centering
  \caption{\textbf{ReVSI \emph{64\_frame} per category} (6{,}616 questions, 380 scenes, 64 frames per scene). Avg is the unweighted mean of the seven categories. All rows are 0.5B. Bold marks the best among our rows.}
  \label{tab:supp-revsi-64frame}
  \small
  \begin{tabular*}{\textwidth}{@{\extracolsep{\fill}} l rrrrrrr r @{}}
    \toprule
    Method & Obj.\ Count & Abs.\ Dist. & Obj.\ Size & Room Size & Rel.\ Dist. & Rel.\ Dir. & Route & Avg \\
    \midrule
    Cambrian-S-0.5B (official, full fp32) & 44.57 & 32.09 & 44.04 & 32.52 & 35.34 & 35.14 & 39.43 & 37.59 \\
    \midrule
    Frames (SigLIP2) & 32.45 & 36.91 & 32.17 & 24.93 & 31.20 & 33.86 & 31.90 & 31.92 \\
    Frames (DINOv3) & 32.09 & 36.02 & 35.60 & 31.36 & 29.63 & 36.56 & 30.47 & 33.10 \\
    Frames + spatial FT (control) & 36.59 & 36.07 & \textbf{48.13} & 45.03 & 32.53 & 37.69 & 44.09 & 40.02 \\
    BEV, SigLIP2 (hard $\times$ FP) & 37.71 & 37.79 & 46.36 & 48.97 & 30.65 & 36.49 & 44.44 & 40.34 \\
    BEV, hard $\times$ world & \textbf{38.61} & 39.58 & 46.89 & 60.86 & 31.20 & 35.06 & 43.73 & 42.27 \\
    \textbf{Cambrian-BEV (hard $\times$ FP)} & 37.65 & 41.72 & 46.31 & 60.23 & 34.64 & 44.29 & \textbf{50.18} & 45.00 \\
    BEV, bilinear $\times$ world & 38.32 & 39.92 & 47.21 & 60.03 & 35.03 & 45.50 & 49.10 & 45.02 \\
    BEV, bilinear $\times$ FP & 37.96 & \textbf{44.53} & 47.21 & \textbf{62.48} & \textbf{37.45} & \textbf{46.17} & 47.67 & \textbf{46.21} \\
    \bottomrule
  \end{tabular*}
\end{table*}

\begin{figure*}[t]
  \centering
  \includegraphics[width=\linewidth]{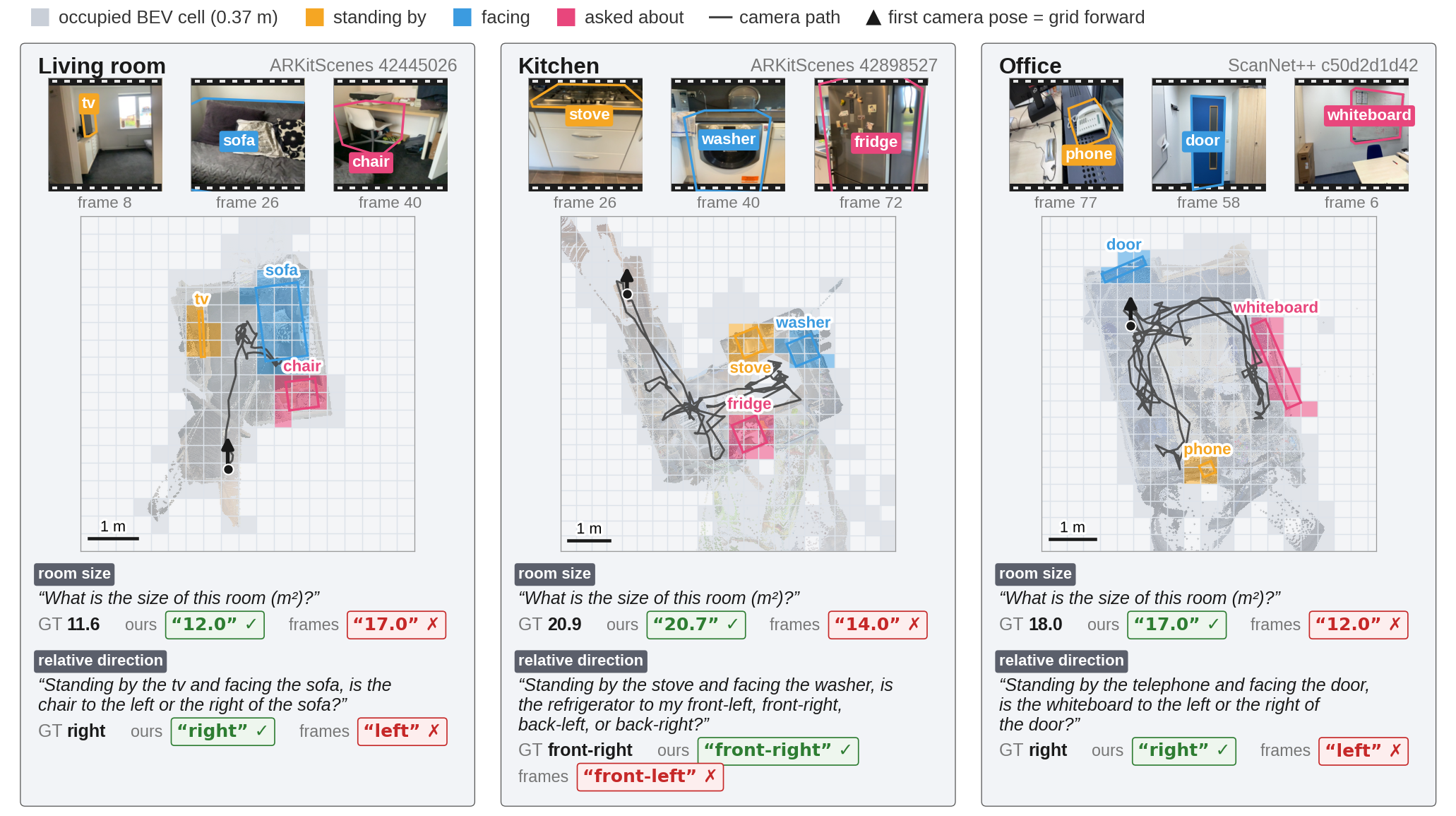}
  \caption{\textbf{BEV occupancy on the floor plan and questions the frame baseline gets wrong.}
  Three VSI-Bench scenes (ARKitScenes living room and kitchen, ScanNet++ office). Top: for each
  object named in the direction question, the video frame in which it is most visible, with the
  dataset's annotated 3D box projected into the image. Middle: top-down rendering of the RGB-D
  reconstruction with the model's $27{\times}27$ grid ($0.37$\,m cells, $\pm5$\,m) in the grid's own
  frame; grey cells receive at least one patch token, the dark line is the camera path, and the
  arrow marks the first camera pose whose heading defines the grid's forward axis. Coloured cells
  and outlines are the same annotated boxes seen from above (orange: standing by, blue: facing,
  pink: asked about). Bottom: the room-size and relative-direction questions from these scenes; the
  frames-only control (``frames'', $9{,}216$ frame tokens) answers them incorrectly and Cambrian-BEV
  (``ours'', 729 BEV tokens) correctly. The occupied area and the egocentric frame of the grid are
  the two quantities behind the largest gains in \cref{tab:main-results} (room size $+24.5$,
  relative direction $+34.1$).}
  \label{fig:supp-qualitative}
\end{figure*}

\begin{figure*}[t]
  \centering
  \includegraphics[width=\linewidth]{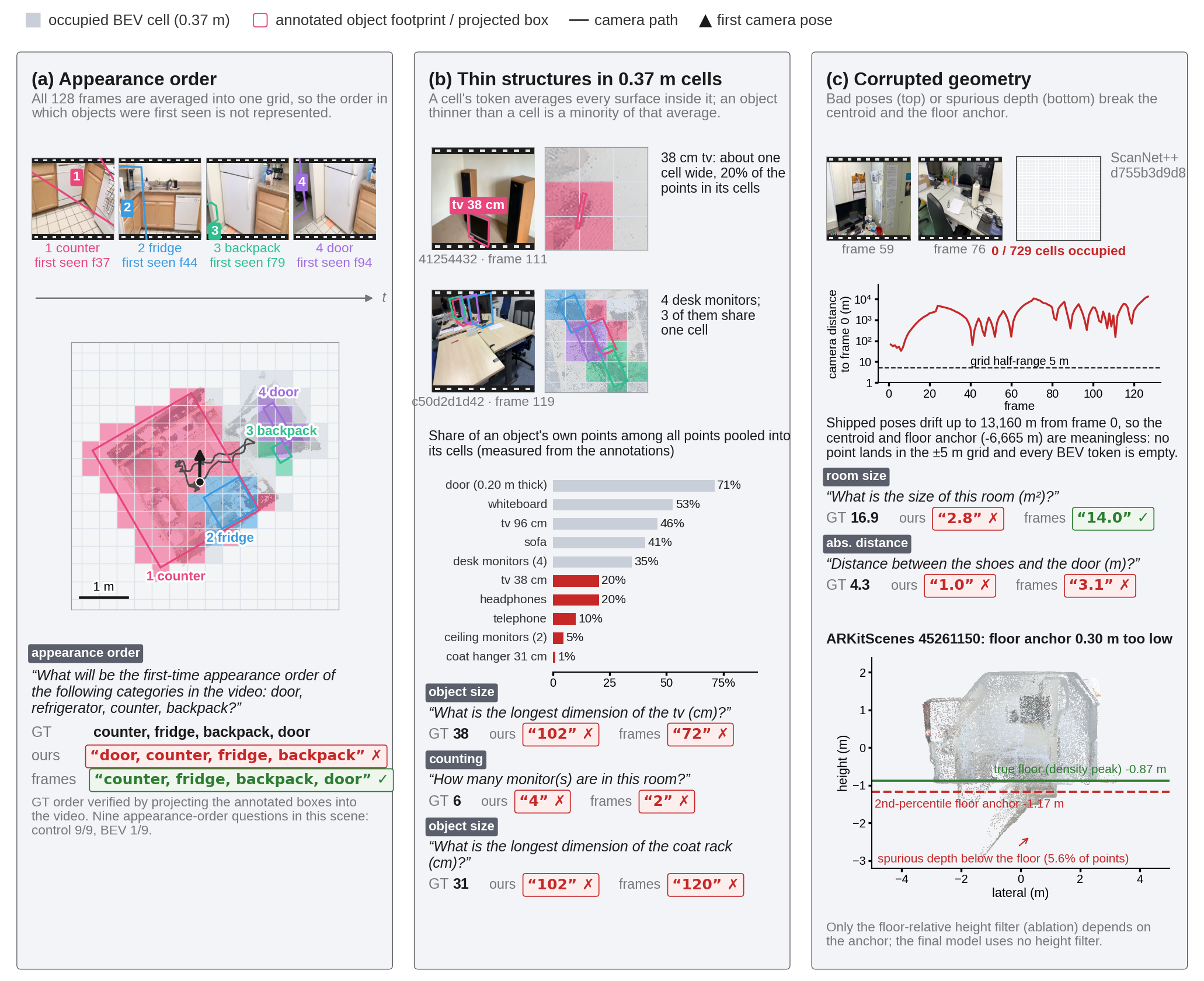}
  \caption{\textbf{Failure cases.} (a) Appearance order (ScanNet \texttt{scene0164\_02}): the
  frames in which each object first becomes visible (annotated boxes projected into the video:
  counter, refrigerator, backpack, door) are all pooled into one grid, so this order is not
  represented; the control answers all nine appearance-order questions of this scene, ours one.
  (b) Thin structures: the token of a cell averages every surface inside it. Measured with the
  datasets' object annotations, a door or a whiteboard contributes 71\% and 53\% of the points pooled
  into its cells, a 38\,cm television 20\%, a telephone 10\% and a 31\,cm coat hanger 1\%; three of
  the four desk monitors in the office share a single cell. Size and count questions about such
  objects fail for both models. (c) Corrupted geometry: in ScanNet++ \texttt{d755b3d9d8} the shipped
  poses drift kilometres from the first frame, so the centroid and the floor anchor are meaningless,
  no point falls inside the $\pm5$\,m grid, every BEV token is empty, and the model answers from its
  prior (room size 2.8 vs.\ 16.9\,m$^2$). In ARKitScenes \texttt{45261150} spurious depth below the
  floor (5.6\% of points) pulls the 2nd-percentile floor anchor 0.30\,m below the true floor; this
  affects only the floor-relative height filter of \cref{sec:supp-height-filter}, which the final
  model does not use.}
  \label{fig:supp-failures}
\end{figure*}

\section{Reproduction gap}
\label{sec:supp-repro-gap}
Our reproduced frame baselines sit $\sim$10 points below the official 0.5B release. The training data are the complete released sets at every stage; the difference is numerical precision. The official recipe trains all four stages in full fp32, whereas we keep fp32 master weights and optimizer states but run the forward and backward passes under bf16 autocast, a choice made to stay within our compute budget: on matched hardware full fp32 is 4.8--6.9$\times$ slower per step at stages 1--2, and slower still at the long-sequence video stages, where fp32 has no fused attention kernel and forces gradient checkpointing; weighted by compute, the four-stage programme would cost roughly 7.5$\times$ more. A controlled stage-2 comparison confirms the direction of the effect (+1.55 on the image-benchmark average for full fp32), and the loss compounds through the later stages (\cref{tab:supp-stages}). We expect our conclusions to carry over to full-precision training and to larger decoders because every comparison in this paper is made \emph{within} a precision and a base model: the frames-only control and the BEV model share the same parent, data, schedule, and precision, so precision shifts both arms together. Moreover, the gains concentrate in room size and relative direction, where our 0.5B model already exceeds the official fp32 7B, and in absolute distance: these answers depend on metric information that frame tokens do not carry at any precision or model size, and the BEV tokens are an input-side representation that leaves the decoder and the four-stage recipe unchanged. The gap only strengthens the comparison: our BEV model starts from the \emph{weaker} base and still surpasses the official release by nearly 12 points.

\section{Patch-level text-similarity probe}
\label{sec:supp-text-sim}

This section specifies how the maps of \cref{fig:text-sim} are produced. The probe asks one question of a frozen encoder: given a text concept, which patch tokens of an image carry evidence for it? It involves only the published encoders. No component of any model we train enters it, so it cannot be confounded by our fine-tuning.

\paragraph{Why a joint-space probe.} Patch tokens are comparable to text only in the space that the encoder's own text tower was trained against. Each encoder is therefore probed with its own text tower and with the patch features that live in that tower's space. Scoring both encoders through a third, shared text encoder would measure that encoder's alignment rather than the locality of the features our BEV grid pools.

\paragraph{Encoders and text towers.} For SigLIP2 we use \path{google/siglip2-so400m-patch14-384}, with the pooled output of its text tower as the text embedding. For DINOv3 we use DINO.txt~\cite{jose2025dinotxt,simeoni2025dinov3} (\path{dinov3_vitl16_dinotxt_tet1280d20h24l}), the text encoder and vision head aligned to the same DINOv3 ViT-L/16 backbone as the models in \cref{tab:main-results}. Its $1280$-d text embedding consists of a CLS-aligned half and a patch-aligned half; we keep the patch-aligned half. Both are the public base checkpoints rather than the towers of our fine-tuned models, because our trained projector maps patch features into the language model's embedding space, to which no text encoder is comparable.

\paragraph{Dense image features.} For DINOv3 the patch features are the projected patch tokens of the DINO.txt vision head, which is what its text tower was trained against. For SigLIP2 the raw \texttt{last\_hidden\_state} patch tokens sit one projection short of the text-aligned space and score as noise per patch. We therefore carry each token through the value pathway of the attention-pooling head, in the manner of MaskCLIP~\cite{zhou2022maskclip}: $v = W_V x$, followed by the head's output projection and residual MLP, $v + \mathrm{mlp}(\mathrm{ln}(v))$. Attention in that head only determines how much of each token's value vector reaches the pooled embedding, so this quantity is the per-patch analogue of what the text tower is trained against. The choice favors SigLIP2: its raw pre-head tokens give maps that are more scattered still.

\paragraph{Preprocessing.} Both encoders see the image shown in the first column of \cref{fig:text-sim}. For this probe the photograph is padded to a square with the processor's mean color (the grey bars) rather than cropped or stretched, so that every patch keeps its aspect ratio; this is the standard Cambrian-S image preprocessing and is specific to the probe. Our BEV pipeline instead resizes each frame anisotropically to $448{\times}448$ and rescales the intrinsics to match (\cref{sec:supp-projection}). The padded image is bilinearly resampled to each tower's native resolution and normalized in that tower's convention, namely $384{\times}384$ with $[-1,1]$ scaling for SigLIP2 and $448{\times}448$ with ImageNet mean and standard deviation for DINOv3. With patch sizes $14$ and $16$ the patch grids are $27{\times}27$ and $28{\times}28$; since the grids differ, the panels are compared on spatial structure rather than cell for cell. Patch features and text embeddings are $L^2$-normalized, so their inner product is a cosine.

\paragraph{Query embedding.} The query is the bare noun in the row label (``tv'', ``bicycle''). In both spaces every text prompt carries a large query-independent component, roughly ``some image content'', and if it is left in place the maps of different queries share most of their spatial structure. We remove it. Let $b$ be the $L^2$-normalized mean embedding of $16$ generic prompts (\emph{a photo, an object, a thing, background, a surface, a room, an image, texture, the scene, an area, material, a place, something, indoor scene, empty space, clutter}); the query embedding $t$ is replaced by $\mathrm{normalize}(t - b\,(t \cdot b))$. In a held-out check this lowered the correlation between the maps of different queries from $0.43$ to $0.29$. The same projection is applied to both encoders, and the contrast in \cref{fig:text-sim} is present with or without it.

\paragraph{Display.} The similarity map is upsampled bilinearly to the image resolution, colored with the \texttt{jet} colormap, and alpha-blended over the image with $\alpha = 0.6$. The color range of each panel is that panel's 2nd and 98th percentile of cosine similarity; values outside are clipped, and the two percentiles are printed at the ends of the panel's colorbar. Clipping is a display choice, not a change to the data: it keeps a few extreme patches from flattening the rest of the map. It is also why each panel has its own colorbar. The two encoders occupy different similarity ranges (\cref{tab:supp-text-sim}), and a single shared bar would misrepresent one of them.

\paragraph{Panel statistics.} \Cref{tab:supp-text-sim} gives, for each panel, the cosine distribution over all patch tokens and Moran's $I$, the spatial autocorrelation of the map over 4-neighbouring patches ($+1$ for smooth regions, $0$ for a spatially random map; shuffling the patches of any of the four maps gives $|I| < 0.002$). The distribution statistics justify the per-panel colorbars: the two encoders differ in mean, floor, and spread, so one shared range would clip one of them. They do not by themselves speak to locality, since they are unchanged under any spatial permutation of the patches. Moran's $I$ does: it is $0.33$ for both SigLIP2 maps and $0.74$ to $0.75$ for both DINOv3 maps. Neighbouring SigLIP2 patches are only weakly correlated, which is the speckle visible in \cref{fig:text-sim}; neighbouring DINOv3 patches agree, which is the compact response over the object.

\paragraph{Images.} The two images are natural photographs from the held-out image validation split used by our visualizer, not VSI-Bench scenes; the probe is run on ordinary photographs on purpose, to show that the property belongs to the encoders rather than to scan data.

\section{ReVSI per-category results}
\label{sec:supp-revsi}
\Cref{tab:supp-revsi-allframe} and \cref{tab:supp-revsi-64frame} give the per-category breakdown behind the averages of \cref{tab:revsi-results}, for the \emph{all\_frame} annotation (128 frames per scene) and the \emph{64\_frame} annotation (64 frames per scene) respectively.

\section{Qualitative examples and failure cases}
\label{sec:supp-qualitative}
\Cref{fig:supp-qualitative} overlays the model's BEV occupancy on top-down renderings of three VSI-Bench scenes together with questions that the identically fine-tuned frames-only control answers incorrectly and Cambrian-BEV answers correctly. The occupied cells trace the metric footprint of the room and the first-pose arrow fixes the egocentric frame, the two quantities behind the room-size and relative-direction gains of \cref{tab:main-results}. The positions of the objects named in each question are taken from the datasets' own 3D annotations (ARKitScenes 3DOD boxes, ScanNet++ instance segments), registered to the benchmark's camera poses, and the ground-truth answer of every question shown was checked against these footprints. \Cref{fig:supp-failures} shows the three characteristic failure modes of the representation: temporal order is discarded by the time-collapsed grid, objects thinner than a $0.37$\,m cell are averaged with the surfaces behind them, and scenes with corrupted poses or depth outliers produce a meaningless centroid and floor anchor. All model answers are copied verbatim from the evaluation logs of the two models.

\end{document}